\documentclass[letterpaper, 10 pt, conference]{ieeeconf}

\IEEEoverridecommandlockouts                              

\usepackage{amsmath} 
\usepackage{amssymb}  
\usepackage{graphicx}
\usepackage{bbm}
\usepackage{xcolor}
\usepackage{hyperref}
\usepackage{subfig}
\usepackage{lipsum}  

\title{\LARGE \bf
Domain Knowledge Driven Pseudo Labels for Interpretable Goal-Conditioned Interactive Trajectory Prediction
}
\author{Lingfeng Sun$^{*1}$, Chen Tang$^{*1}$, Yaru Niu$^{2}$, Enna Sachdeva$^{3}$, \\ Chiho Choi$^{3}$, Teruhisa Misu$^{3}$, Masayoshi Tomizuka$^{1}$,  Wei Zhan$^{1}$
\thanks{$^{*}$ Equal contribution.}
\thanks{$^{1}$ Department of Mechanical Engineering, University of California Berkeley, CA, USA}
\thanks{$^{2}$ School of Electrical and Computer Engineering, Georgia Institute of Technology, GA, USA}
\thanks{$^{3}$ Honda Research Institute, CA, USA}
\thanks{Corresponding author: Chen Tang (email: chen\_tang@berkeley.edu)}
}

\begin{document}

\maketitle
\begin{abstract}
Motion forecasting in highly interactive scenarios is a challenging problem in autonomous driving. In such scenarios, we need to accurately predict the joint behavior of interacting agents to ensure the safe and efficient navigation of autonomous vehicles. Recently, goal-conditioned methods have gained increasing attention due to their advantage in performance and their ability to capture the multimodality in trajectory distribution. In this work, we study the joint trajectory prediction problem with the goal-conditioned framework. In particular, we introduce a conditional-variational-autoencoder-based (CVAE) model to explicitly encode different interaction modes into the latent space. However, we discover that the vanilla model suffers from posterior collapse and cannot induce an informative latent space as desired. To address these issues, we propose a novel approach to avoid KL vanishing and induce an interpretable interactive latent space with pseudo labels. The proposed pseudo labels allow us to incorporate domain knowledge on interaction in a flexible manner. We motivate the proposed method using an illustrative toy example. In addition, we validate our framework on the Waymo Open Motion Dataset with both quantitative and qualitative evaluations.
\end{abstract}
\section{Introduction} \label{sec:introduction}
\begin{figure}[t]
    \centering
    \subfloat[\centering]{{\includegraphics[width=1.4in]{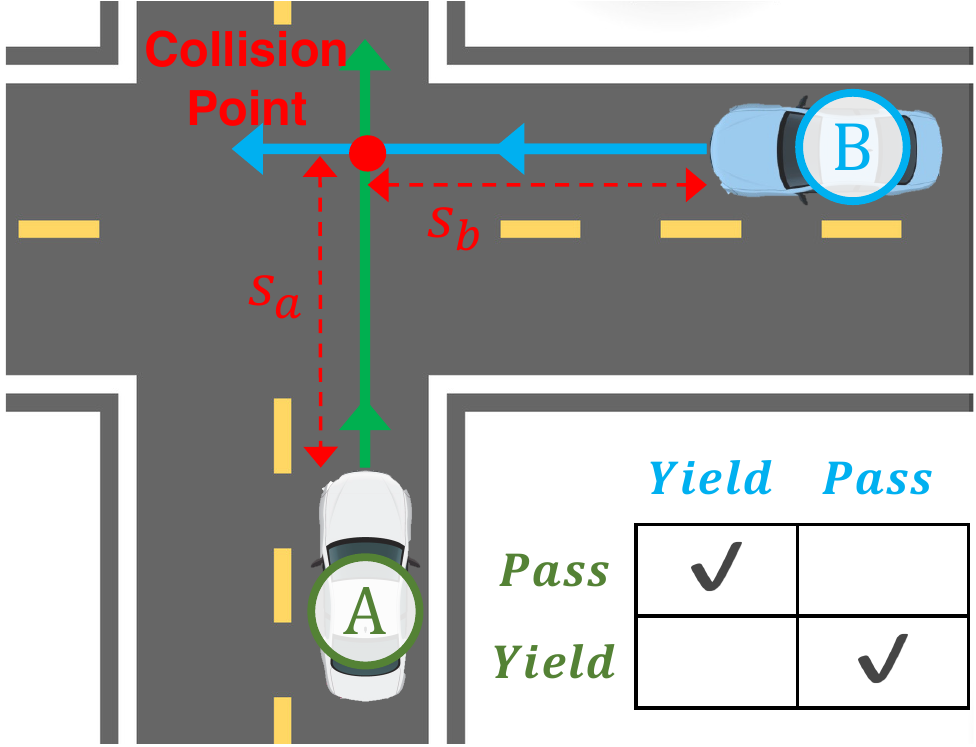}}\label{fig:toy_example}}
    \quad
    \subfloat[\centering]{{\includegraphics[width=1.3in]{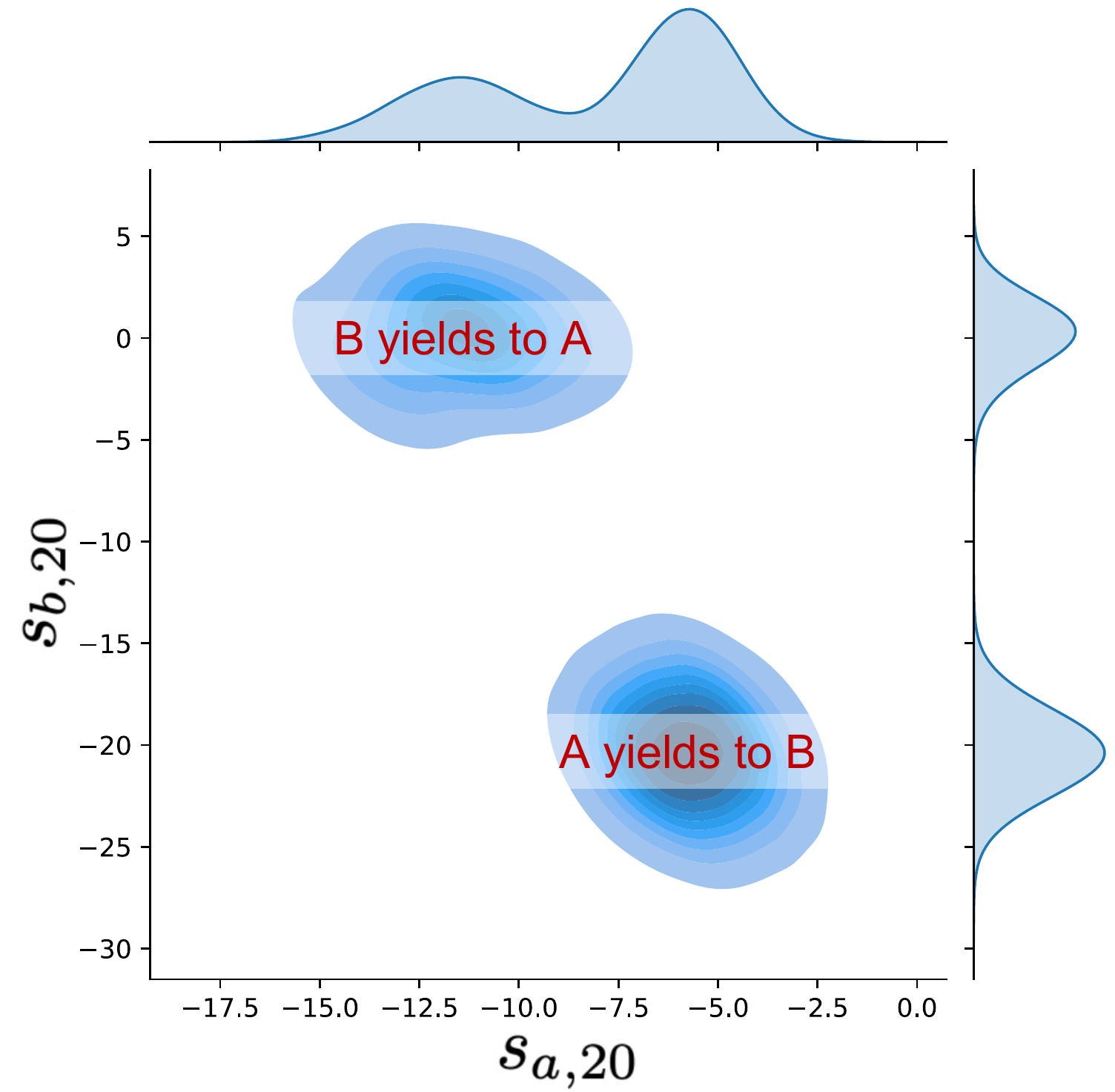}} \label{fig:marginal_vs_joint}}
    \caption{A motivating toy example. (a) depicts the scenario of the toy example, where two cars are driving towards a collision point at an intersection. (b) shows the ground-truth joint distribution and marginal distribution of $s_{a}$, $s_{b}$ at the 20$^\textrm{th}$ time step when the behavior of the cars is govern by the model described in Sec.~\ref{sec:toy_example}.}
    \vspace{-0.2in}
\end{figure}

Autonomous vehicles need to accurately predict other road participants' behaviors to navigate safely and efficiently in complex driving scenarios. Previous prediction benchmarks mainly focus on single-agent settings~\cite{interactiondataset}. When multiple agents exist, the predicted trajectories are evaluated independently for each agent. Consequently, predicting the \emph{marginal distribution} of vehicle trajectories suffices for achieving good results on those benchmarks. However, such models may generate unrealistic predictions in highly interactive scenarios. For example, at the intersection illustrated in Fig.~\ref{fig:toy_example}, each vehicle has two possible motion patterns: entering the intersection and yielding before the intersection. A model predicting the marginal distributions might predict infeasible joint behaviors (i.e., both cars follow the same motion pattern). To accurately assess such kinds of interactive behaviors, it is then necessary to predict the \emph{joint distribution} of the interacting agents' future trajectories (Fig.~\ref{fig:marginal_vs_joint}). Recently, Waymo provided an interaction prediction benchmark based on the Waymo Open Motion Dataset (WOMD)~\cite{Ettinger_2021_ICCV}, where the trajectories of two interacting agents are predicted and evaluated jointly. It serves as an ideal test bed and motivates us to study the interaction prediction problem. 

We are particularly interested in the interaction prediction problem under the goal-conditioned framework, as goal-conditioned methods can effectively capture the multimodality in trajectory distribution \cite{mangalam2020not, zhao2020tnt, gu2021densetnt_iccv}. Under this framework, we first explicitly predict the distribution of an agent’s endpoint over a discretized goal set and then complete the trajectories conditioned on the selected goal points. However, previous methods mainly focus on single-agent prediction. For multiple agents, these methods predict the trajectories independently for each agent. To model the joint distribution of interacting agents’ goals, we extend the goal set to a goal-pair set which allows joint prediction of two agents’ endpoints. By choosing a dense set as in \cite{gu2021densetnt_iccv}, this categorical distribution of goal pair can reasonably approximate the joint distribution in any interactive scenarios.

In practice, downstream modules require a small set of representative predictions \cite{deo2020trajectory}. The limited onboard computational resource also restricts the number of sampled trajectories. For the downstream module to understand the interactive scenario precisely, it becomes critical to ensure that different interaction modes can be efficiently captured with a limited number of sampled trajectories. To this end, we leverage the Conditional Variational Autoencoder (CVAE) framework~\cite{sohn2015learning, li2019conditional, tang2021exploring} and introduce a discrete latent space to capture the interaction modes explicitly~\cite{ivanovic2019trajectron, salzmann2020trajectron++}. For instance, in the toy example, we want different latent variables to represent different right-of-ways, corresponding to the two modalities in the joint distribution of goal pairs shown in Fig.~\ref{fig:marginal_vs_joint}. However, it is not guaranteed that the model can always learn an informative latent space distinguishing interaction modes useful for downstream modules.

In our goal-conditioned CVAE-based framework, the goal pair follows a categorical distribution. It changes the reconstruction task into a multi-label classification problem. Without knowing the distance between the goal pairs, we find it difficult for the model to distinguish between them. Therefore, it becomes difficult to determine which goal pairs should be encoded into the same latent variable, which leads to the problem of posterior collapse in CVAE, resulting in an uninformative latent space.

To tackle this problem, we propose to guide the training with \emph{pseudo labels}\footnote{For clarification, we refer any generated labels other than the ground-truth ones as pseudo labels. They are not necessarily generated for the purpose of semi-supervised learning or self-supervised learning.} designed based on domain knowledge. For each ground-truth goal pair, we assign positive target values to goal pair candidates similar to it. The model learns to encode similar goal pairs into the same latent variable by minimizing the distance between the decoded distribution and the pseudo labels. Since the goal pair distribution is defined over a fixed finite set, the pseudo labels can be pre-computed for each goal pair candidate. Therefore, we do not require the computation of pseudo labels to be differentiable. It allows us to incorporate domain knowledge into the pseudo labels in a flexible manner and specify any interaction modes for the latent space to capture. 

Our contributions are two-fold: \emph{(1)} We present a goal-conditioned CVAE model for the joint trajectory prediction task of interacting pairs. \emph{(2)} We propose a novel and flexible approach to induce an interpretable interactive latent space using pseudo labels. In particular, we introduce three types of pseudo labels corresponding to different domain knowledge on interaction. We show that the proposed pseudo labels can effectively enforce an interpretable latent space in an illustrative toy example and on real-world traffic datasets.

\section{Related Works}
{\bf Interactive Trajectory Prediction.} Most of prior efforts in trajectory forecasting in interactive traffic scenarios focused on modeling interaction in input representations. Different model architectures were proposed to encode social context from the historical trajectories of interacting agents and HD maps~\cite{alahi2016social, mercat2020multi, gao2020vectornet, liang2020learning}. As argued in Sec.~\ref{sec:introduction}, it is necessary to model the joint behavior of the target agents in the prediction for accurate forecasting in highly interactive scenarios. Some works applied regularization to discourage unrealistic joint behavior such as collision in the predicted outcomes~\cite{tolstaya2021identifying, suo2021trafficsim}. However, such kinds of heuristic approaches may introduce bias into the model. For instance, the resulting models cannot detect traffic accidents if collisions are ruled out in the predicted trajectories. Other works attempted to directly approximate the joint trajectory distribution from the dataset by employing step-wise multi-agent roll-outs~\cite{tang2019multiple}, directly predicting a fixed number of multiple joint trajectories~\cite{ngiam2021scene}, or factorizing the joint distribution by classifying agents as influencers and reactors in interacting pairs~\cite{sun2022m2i}. In this work, we approximate the joint trajectory distribution of agent pairs via modeling the distribution of their goals defined over a discrete set of goal pairs, which enables the usage of pseudo labels as a flexible manner to incorporate domain knowledge on interaction into the prediction model. 

\section{Problem Formulation}
\subsection{Background: Goal-Conditioned Prediction} \label{sec:prelim-goal}
In general, a trajectory prediction model learns to model the distribution $p(\boldsymbol{y}|\boldsymbol{T})$, where $\boldsymbol{y}$ denotes the future trajectory of the target agent, and $\boldsymbol{T}$ denotes the embedding of the agent's history and context information. In goal-conditioned trajectory prediction framework, the prediction task consists of two stages: goal prediction and trajectory completion, resulting in the decomposition of $p(\boldsymbol{y}|\boldsymbol{T})$:
\begin{equation*}
    p(\boldsymbol{y}|\boldsymbol{T})=\int_{g\in \mathcal{G}}p(\boldsymbol{y}|g, \boldsymbol{T})\cdot p(g|\boldsymbol{T})dg,
\end{equation*}
where $\mathcal{G}$ is the goal space. The goal-prediction model $p(\boldsymbol{g}|\boldsymbol{T})$ can capture the multi-modality in driver intention, while the goal-conditioned trajectory completion module models the driving behavior to reach the goals. 

The overall framework has three stages. The first stage is \emph{goal distribution prediction}. Depending on the goal space, $p(\boldsymbol{g}|\boldsymbol{T})$ can be modeled as either a continuous or discrete distribution. We are particularly interested in the formulation of \cite{gu2021densetnt_iccv}, in which $\mathcal{G}$ is defined as a dense and discretized goal set covering the drivable area. In such a way, $p(\boldsymbol{g}|\boldsymbol{T})$ directly models the distribution of goal points, instead of anchor points as in \cite{zhao2020tnt}. The second stage is \emph{goal-conditioned trajectory prediction}, where the conditional distribution of future motions is modeled as a simple unimodal distribution (e.g., Gaussian distribution). The third stage is \emph{sampling and selecting}, where a final small number of predictions are selected to fulfill the requirement of downstream applications. The commonly used techniques are heuristic-based algorithms, such as non-maximum suppression (NMS) \cite{zhao2020tnt}.

\subsection{Goal-conditioned Interactive Prediction}
The framework described in Sec. \ref{sec:prelim-goal} is primarily designed for single-agent prediction. The extension of this two-stage prediction scheme to multi-agent settings is not straightforward. In multi-agent trajectory prediction, we need to model the joint distribution of all agents' future trajectories, i.e., $p\left(\boldsymbol{y}_1, \boldsymbol{y}_2, \cdots, \boldsymbol{y}_N|\boldsymbol{T}\right)$. We can decompose the interacting agents at the trajectory completion stage by adopting the assumption that the trajectories are independent after conditioning on the goals. However, we still need to model the joint distribution of their goals, i.e., $p(\boldsymbol{g}_1, \boldsymbol{g}_2,\cdots,\boldsymbol{g}_N|\boldsymbol{T})$. We cannot simply assume the trajectories of interacting agents are independent and decompose the joint distribution into $\prod_{i=1}^N p(\boldsymbol{g}_i|\boldsymbol{T})$. The simplified distribution cannot model the interactive behavior between agents, for instance, the fundamental interacting rule\textemdash collision avoidance. Meanwhile, if we directly model the joint distribution, we need to select a discrete goal set $\mathcal{G}_i$ for each modeled agent $i$. The overall dimension of the joint distribution becomes $\prod_{i=1}^N|\mathcal{G}_i|$, which grows exponentially with the number of agents. 

To mitigate the curse of dimensionality, we first predict the marginal distributions of the goals. Afterward, we use the marginal distributions to prune the goal sets $\{\mathcal{G}_i\}_{i=1}^N$. Concretely, we select $M$ goal candidates with the highest marginal probability for each agent. In our experiments, we find that we can reasonably approximate the marginal distribution with $M<<|\mathcal{G}_i|$. It is then sufficient to model the distribution of $|M|^N$ goal combinations, which is applicable for the prediction task of interacting pairs. 

\section{Inducing Interpretable Interactive Latent Space with Pseudo Labels} \label{sec:toy_example}
In this section, we take the scenario illustrated in Fig.~\ref{fig:toy_example} as a running example to introduce the proposed pseudo labels. Specifically, we explain the motivation and demonstrate how the pseudo labels may help induce an interpretable interactive latent space in this toy example. 

As shown in Fig.~\ref{fig:toy_example}, Vehicle $A$ and $B$ are driving towards a collision point. The states of the vehicles are $s_a, s_b$ and $v_a, v_b$, where $s_{a, b}$ are the displacements of the vehicles $A, B$ relative to the collision point and $v_{a, b}$ are the absolute velocities. Each vehicle is assigned a target position to follow at each step, depending on which vehicle has the ``right-of-way''. If a vehicle has the right-of-way, we assign a point that is substantially far away along its driving direction as its target point. If the other vehicle has the right-of-way and has not passed the collision point, we assign the collision point as the target point. We assume that the right-of-way is affected by the difference of time headway at the initial time since the car with a shorter headway time to the collision point is more likely to get the right-of-way in interaction. The time headway at timestep $t$ is defined as $T_{\textrm{head}, t} = \max\left(\frac{s_{t}}{v_{t}}, 0\right)$. The probability of Vehicle A getting the right-of-way is set as:
\begin{equation*}
    p_{A} = 0.5\left(\tanh\frac{T_{a,\textrm{head}, 0} - T_{b, \textrm{head}, 0}}{\eta}+1\right),
\end{equation*}
where $\eta$ controls the rate of transition between entering into the intersection and yielding. The dynamics of the vehicles is governed by the intelligent driver model \cite{treiber2013traffic}. Each vehicle follows the target position set according to its right-of-way.

The task is to jointly predict the endpoints $\boldsymbol{g}=(\boldsymbol{g}_a, \boldsymbol{g}_b)=(s_{a, 20}, s_{b, 20})$ of both vehicles after 20 timesteps, given the initial condition $\boldsymbol{T}=(\boldsymbol{T}_a, \boldsymbol{T}_b)=([s_{a, 0}, v_{a, 0}],[s_{b, 0}, v_{b, 0}])$. It is analogous to the goal prediction stage in the goal-conditioned prediction framework. The joint goal distribution is defined over a discrete set $\mathcal{G}_{a,b}$ obtained by discretizing the spaces of $s_{a,20}$ and $s_{b,20}$. Given the same initial conditions, there are two interaction modes, i.e., Vehicle A yields to Vehicle B and vice versa. These two interaction modes result in a multi-modal joint goal distribution as shown in Fig.~\ref{fig:marginal_vs_joint}. 

\begin{figure*}[t]
    \centering
    \subfloat[\centering Vanilla CVAE Model]{{\includegraphics[width=5.9cm]{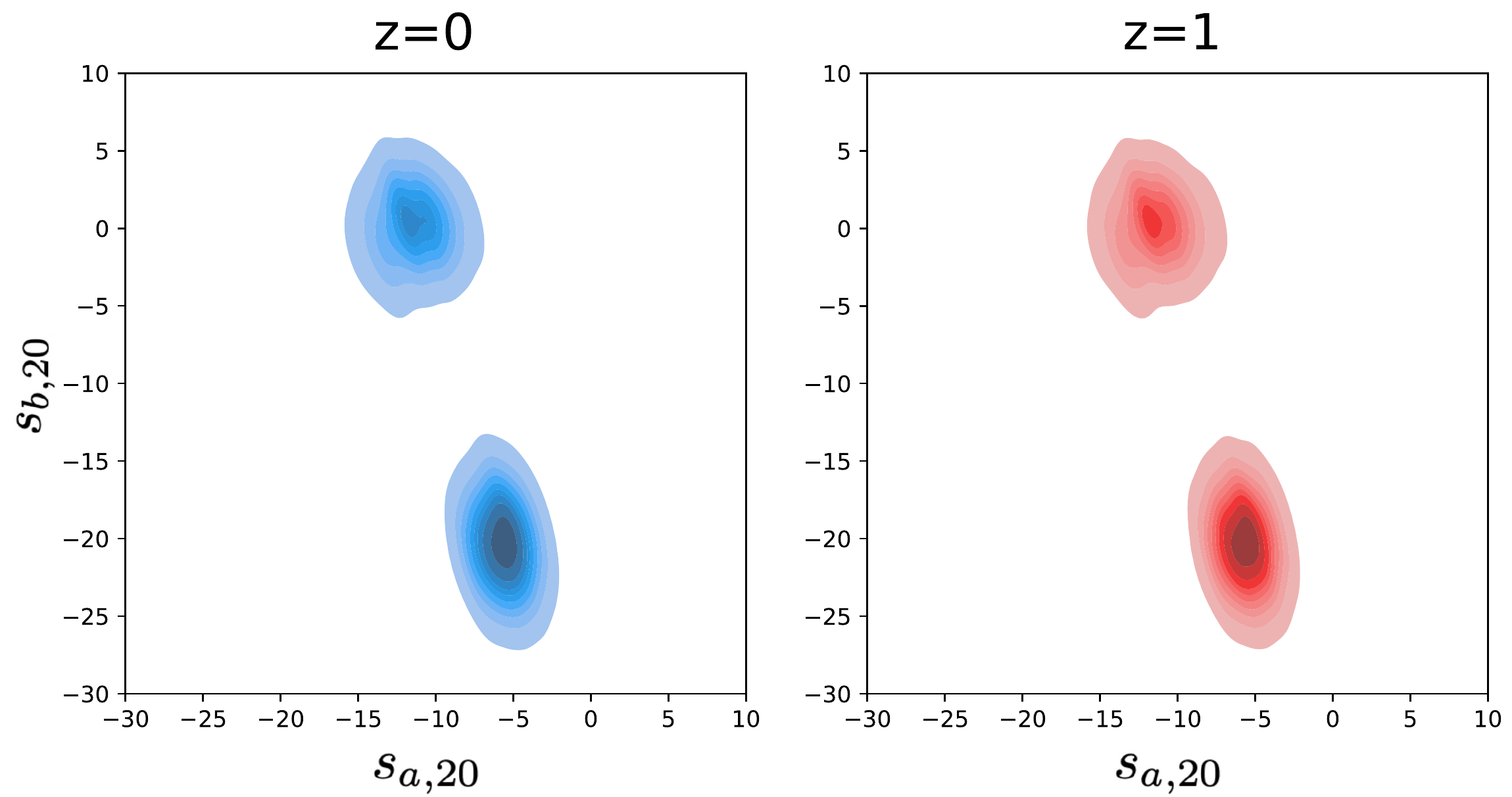} }\label{fig:toy_joint_vanilla_vae}}
    \qquad
    \subfloat[\centering CVAE Model with Pseudo Distance Labels]{{\includegraphics[width=6.2cm]{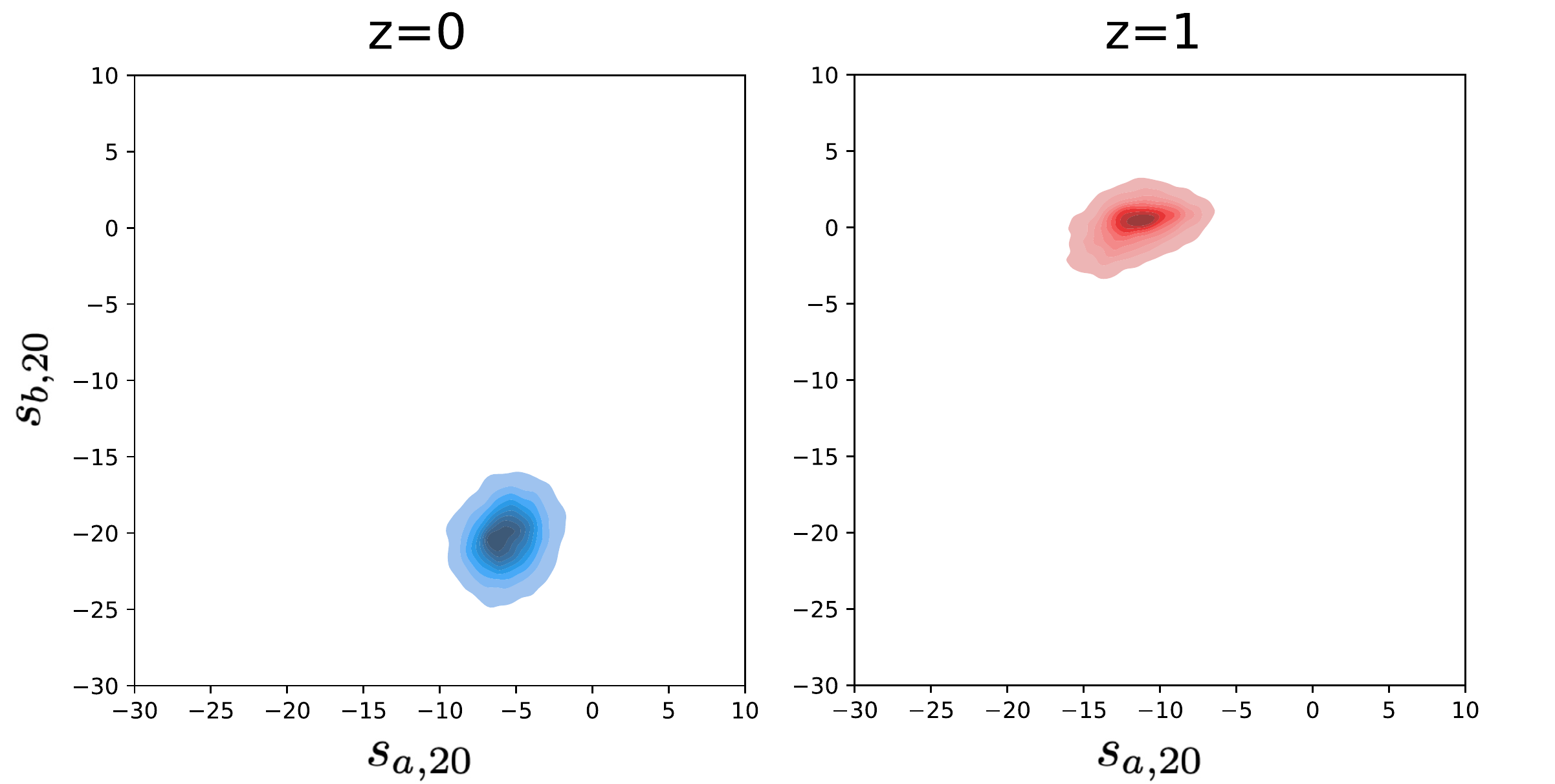} \label{fig:toy_joint_distance_vae}}}
    \caption{Joint goal distributions decoded from different latent variables with different models. With the pseudo distance labels, the model is able to capture the two modes in its latent space. The results are the same when using other pseudo labels.}
    \vspace{-0.1in}
    \label{fig:toy_joint_vae}
\end{figure*}

To model the joint goal distribution, we leverage the CVAE framework with a discrete latent space. The CVAE model consists of three modules: 1) An encoder $q_\theta(\boldsymbol{z}|\boldsymbol{T}, \boldsymbol{g})$ approximating the posterior distribution of $\boldsymbol{z}$; 2) A conditional prior $p_\phi(\boldsymbol{z}|\boldsymbol{T})$; 3) A decoder $p_\psi(\boldsymbol{g}|\boldsymbol{T}, \boldsymbol{z})$ modeling the conditional joint goal distribution. We use MLPs for all the modules. The model is trained by maximizing the evidence lower bound (ELBO):
\begin{align}
     \mathcal{L}({\theta, \phi, \psi}) =& \ -\mathbb{E}_{\boldsymbol{T}, \boldsymbol{g},\boldsymbol{y}\sim \mathcal{D}}\Big\{\mathbb{E}_{\boldsymbol{z}\sim q_\theta\left(\boldsymbol{z}|\boldsymbol{T}, \boldsymbol{g}\right)}\left[f\left(\boldsymbol{y}, p_\psi\left(\cdot|\boldsymbol{T}, \boldsymbol{z}\right)\right)\right] \nonumber  \\
     & - \beta D_{KL}\left[q_\theta(\boldsymbol{z}|\boldsymbol{T}, \boldsymbol{g})\|p_\phi(\boldsymbol{z}|\boldsymbol{T}) \right]\Big\}, \label{eq:elbo}
\end{align}
where $\mathcal{D}$ is the dataset consisting of initial states $\boldsymbol{T}$, goal pairs $\boldsymbol{g}$, and ground-truth labels $\boldsymbol{y}$. The vector $\boldsymbol{y}\in\left\{0, 1\right\}^{|\mathcal{G}_{a, b}|}$ collects ground-truth scores of the goal pairs in $\mathcal{G}_{a,b}$. We assign one to the ground-truth goal pair and zero to the others. We choose the Binary Cross-Entropy (BCE) loss as the function $f$ to define the reconstruction loss. 

\subsection{Avoiding KL Vanishing with Pseudo Labels} 
Our experiments with the CVAE model formulated above show that the KL divergence tends to vanish, and the conditional prior distribution always concentrates into a single value. As shown in Fig. \ref{fig:toy_joint_vanilla_vae}, the latent space is completely uninformative. While the decoder can still model the joint distribution, the model does not fulfill our objective to capture interaction modes with the latent space explicitly. This phenomenon is similar to the posterior collapse problem that occurs when an autoregressive decoder is used in sequence modeling \cite{fu2019cyclical}. The MLP decoder we use can model the joint distribution without the latent space. With such a powerful decoder, the model is prone to ignoring the latent space to minimize the KL divergence. 

We find it difficult for the model to escape from posterior collapse in our case. To gain some insights into the reason behind, consider a special case where the dataset is collected under the same conditions and the weight of KL regularization $\beta$ equals to zero. A zero $\beta$ value occurs during the training procedure when KL annealing~\cite{fu2019cyclical} is applied to mitigate KL vanishing. We will show that it is still difficult to prevent posterior collapse even if we set $\beta=0$. In this case, the optimal posterior distribution always assigns all the probability mass to a single latent variable. Consequently, we can consider the VAE as solving a clustering problem. Given a $d_z$-dimensional discrete latent space, the VAE model essentially clusters $\mathcal{G}_{a,b}$ into $d_z$ subgroups, denoted as $\left\{S_k\right\}^{d_z}_{k=1}$, and finds a distribution of goal pairs minimizing the BCE loss for each subgroup. We can easily obtain the minimal value of the BCE loss within a given subgroup analytically. It is then straightforward to see that the optimal clustering scheme essentially minimizes the sum of the objectives over the subgroups, which is the following objective function: 
\begin{equation*}
    \mathcal{L}(\left\{S_k\right\}) = \sum_{k=1}^{d_z}\sum_{j\in{S_k}}(n_j-n_{S_k})\log(1 - \frac{n_j}{n_{S_k}})-n_j\log(\frac{n_j}{n_{S_k}}),
    \label{eqn:opt-vae}
\end{equation*}
where we define $n_j$ and $n_{S_i}$ as:
\begin{equation*}
    n_j = \sum_{i=1}^{|\mathcal{D}|} \mathbf{1}\left(y^i_j=1\right), \quad
    n_{S_k} = \sum_{j\in S_k} n_j.
\end{equation*}
In other words, $n_j$ counts how many times the $j^\mathrm{th}$ goal pair appears in the dataset, and $n_{S_k}$ counts how many times the goal pairs in the subgroup $S_k$ appear in the dataset. 

With posterior collapse, the clustering scheme corresponds to having all the elements in a single subgroup while leaving the rest empty. In our toy example, it is easy to check that better solutions do exist, for instance, the one shown in Fig.~\ref{fig:toy_joint_distance_vae}. Since the two modes in the joint distribution are separated in the latent space, the goal pairs have a higher likelihood under the decoded distribution conditioned on the latent variable it corresponds to, which leads to a smaller reconstruction error than the trivial solution resulting from posterior collapse. However, it is difficult for the model to escape from the suboptimal solution shown in Fig.~\ref{fig:toy_joint_vanilla_vae}. The objective $\mathcal{L}(\left\{S_k\right\})$ purely relies on the frequencies of different goal pairs in the dataset. We can interchange goal pairs that appears with similar numbers of times without affecting the objective. It is then difficult for the model to learn which two goal pairs should be assigned to the same latent variable to minimize the objective value. 

To mitigate this issue, we propose to inform the model of the proximity between goal pairs via pseudo labels. For each goal pair $\boldsymbol{g}_j\in \mathcal{G}_{a,b}$, a pseudo label is a vector defined over $\mathcal{G}_{a,b}$ with values ranging from zero to one, which we denoted as $\boldsymbol{\hat{y}}_j\in\left[0, 1\right]^{|\mathcal{G}_{a,b}|}$. In $\boldsymbol{\hat{y}}_j$, we assign positive values to the ground-truth goal pair as well as those goal pairs that are ``close" to $\boldsymbol{g}_j$ by the distance metric defined by domain knowledge, in contrast to the label from the dataset where the positive value is only assigned to the single ground-truth goal pair. We use the pseudo labels to define the following auxiliary loss function: 
\begin{equation*}
    \alpha \mathbb{E}_{\boldsymbol{T}, \boldsymbol{g},\boldsymbol{y}\sim \mathcal{D}, \boldsymbol{z}\sim q_\theta\left(\boldsymbol{z}|\boldsymbol{T}, \boldsymbol{g}\right)}
    \sum_{j=1}^{|\mathcal{G}_{a,b}|} \mathbf{1}(y_{j}=1)f\left(\boldsymbol{\hat{y}}_j, p_\phi(\cdot|\boldsymbol{T},\boldsymbol{z})\right),
\end{equation*}
where the function $f$ quantifies the distance between the pseudo labels and the conditional joint goal distribution. By minimizing the auxiliary loss, the model learns to assign high probabilities to both the ground-truth goal pair and those ``close'' ones specified by the pseudo labels in the distribution conditioned on the same latent variable. Consequently, the model is guided to encode goal pairs that are close to each other into the same latent variable, which prevents the latent space from being totally uninformative. It is worth noting that since the pseudo labels are not required to be generated in a differentiable way, it allows us to flexibly design pseudo labels based on domain knowledge on the proximity between goal pairs. In the next subsection, we will introduce three types of pseudo labels we design in this work. 

\subsection{Pseudo Labels}
\subsubsection{Pseudo Distance Labels}
Since the agents move continuously, their behaviors should be consistent if targeting goal pairs that are close to each other in terms of Euclidean distance. Such goal pairs should then be clustered into the same group. Consequently, we introduce the pseudo distance labels defined as:
\begin{equation*}
    \boldsymbol{\hat{y}}^{\textrm{distance}}_{j, i} = \exp\left(-\frac{\|\boldsymbol{g}_j - \boldsymbol{g}_i\|^2}{2\sigma^2}\right),\ i=1,2,\cdots, d.
\end{equation*}
It essentially smooths the original singular label with the radial basis (RBF) kernel. We choose $f$ as the BCE loss.

 With the auxiliary loss induced by the pseudo distance labels, the CVAE model learns to separate the two interaction modes in the latent space (Fig. \ref{fig:toy_joint_distance_vae}). Also, the prior probabilities of the two latent variables are consistent with the ground-truth probabilities of the corresponding interaction modes in the simulation. The interaction modes can be effectively separated because the Euclidean distance between goal pairs from different clusters is far away. 

\subsubsection{Pseudo Marginal Labels}
The joint goal distribution is the consequence of the interaction between agents. If Agent A targets the same goal regardless of what goal Agent B follows, we may characterize the interaction by the goal of Agent A. Therefore, we consider goal pairs that share the same goal of one agent closer than those that are totally different. We then define two sets of pseudo marginal labels:
\begin{equation*}
    \boldsymbol{\hat{y}}^{\textrm{marginal}, a}_{j, i} = \mathbf{1}\left(\boldsymbol{g}_{j,a}=\boldsymbol{g}_{i,a}\right),\quad 
    \boldsymbol{\hat{y}}^{\textrm{marginal}, b}_{j, i} = \mathbf{1}\left(\boldsymbol{g}_{j,b}=\boldsymbol{g}_{i,b}\right),
\end{equation*}
and the corresponding loss function:
\begin{equation*}
\begin{aligned}
    & f^\textrm{marginal}\left(\boldsymbol{\hat{y}}^{\textrm{marginal}, a}_{j}, \boldsymbol{\hat{y}}^{\textrm{marginal}, b}_{j}, p_\phi(\cdot|\boldsymbol{T},\boldsymbol{z})\right) \\
    = & \log \left(\sum_{i=1}^{|\mathcal{G}_{a,b}|} \mathbf{1}\left(\boldsymbol{\hat{y}}^{\textrm{marginal}, a}_{j,i}=1\right) p_\phi(\boldsymbol{g}_i|\boldsymbol{T},\boldsymbol{z})\right)\\
    + & \log \left(\sum_{i=1}^{|\mathcal{G}_{a,b}|} \mathbf{1}\left(\boldsymbol{\hat{y}}^{\textrm{marginal}, b}_{j,i}=1\right) p_\phi(\boldsymbol{g}_i|\boldsymbol{T},\boldsymbol{z})\right).
\end{aligned}
\end{equation*}
We essentially maximize the log likelihood of the ground-truth goal pairs under the marginal goal distributions. 

With the pseudo marginal labels, we can guide the CVAE model to perfectly separate the goal pairs into two interaction modes in the toy example. The result is the same as shown in Fig.~\ref{fig:toy_joint_distance_vae}. The interaction modes can be perfectly identified because the goal pairs from different clusters happen to have distinct coordinates in both dimensions in our toy example. If only one of the agents changes his behavior in different modes, the pseudo marginal labels alone will not be helpful. 

\subsubsection{Pseudo Interaction Labels}
\label{sec:pseudo-interaction}
The last type of pseudo labels we introduce allows us to incorporate domain knowledge on interaction in a flexible way, which we refer to as pseudo interaction labels. From the perspective of the downstream planner, we may want the latent space to distinguish specific interaction modes for efficient planning and risk evaluation (e.g., collision vs. no collision, yielding vs. passing). If we know that these interaction modes can be identified with certain features, we can design the corresponding pseudo interaction labels as follows:
\begin{equation*}
    \boldsymbol{\hat{y}}^\textrm{interact}_{j,i}\left(\boldsymbol{T}\right) = \mathbf{1}\left(h(\boldsymbol{T}, \boldsymbol{g}_i)=h(\boldsymbol{T},\boldsymbol{g}_j)\right),
\end{equation*}
where the function $h$ maps the goal pair and initial states to a vector of discrete variables characterizing the interaction. We assign positive values to those goal pairs that have the same features as the ground-truth goal pair. It indicates that they are under the same interaction mode as the ground-truth one. Regarding the loss function, maximizing the log likelihood of positive goal pairs could be misleading. There could be a large ratio of goal pair candidates under the same interaction mode. Inspired by \cite{kim2019nlnl}, we adopt a loss function to minimize the probabilities of negative labels:
\begin{equation*}
\begin{aligned}
    & f^\textrm{interact}\left(\boldsymbol{\hat{y}}^\textrm{interact}_{j}, p_\phi(\cdot|\boldsymbol{T},\boldsymbol{z})\right) \\
    = &\sum_{i=1}^{|\mathcal{G}_{a,b}|}  \mathbf{1}(\boldsymbol{\hat{y}}^\textrm{interact}_{j,i}=0)\log\left(1-p_\phi(\boldsymbol{g}_i|\boldsymbol{T},\boldsymbol{z})\right).    
\end{aligned}
\end{equation*}
In the toy example, we adopt an interaction feature indicating which agent has longer displacement in 20 steps, i.e., $\mathbf{1}\left(s_{a,0} - s_{a,20}>s_{b,0} - s_{b, 20}\right)$. With this feature, we can identify which agent decides to yield. By incorporating this pseudo interaction label, we are able to separate the interaction modes in the latent space and obtain a model similar to the one shown in Fig.~\ref{fig:toy_joint_distance_vae}. 

It is worth noting that pseudo interaction labels are only applied to the distribution decoded from the latent variable which the ground-truth goal pair belongs to. In other words, we only require there exists an interaction mode in the latent space that is consistent with the ground-truth, instead of enforcing all the predicted goal pairs to satisfy the constraints. As a result, we can avoid over-regularization and unnecessary bias. Also, we do not require a comprehensive set of pseudo interaction labels covering all kinds of interactive traffic scenes. We can apply the pseudo labels designed for the specific scenarios of interest without worrying about harming the model performance on the other scenarios. For those scenarios where the designed labels are not applicable, all the goal pair candidates have the same features as the ground-truth one. Therefore, the auxiliary loss is always zero and the pseudo labels are simply ignored. 

\section{Framework Architecture}
\begin{figure*}[t]
    \centering
    \subfloat[\centering Overall Model Architecture]{{\includegraphics[width=3.4in]{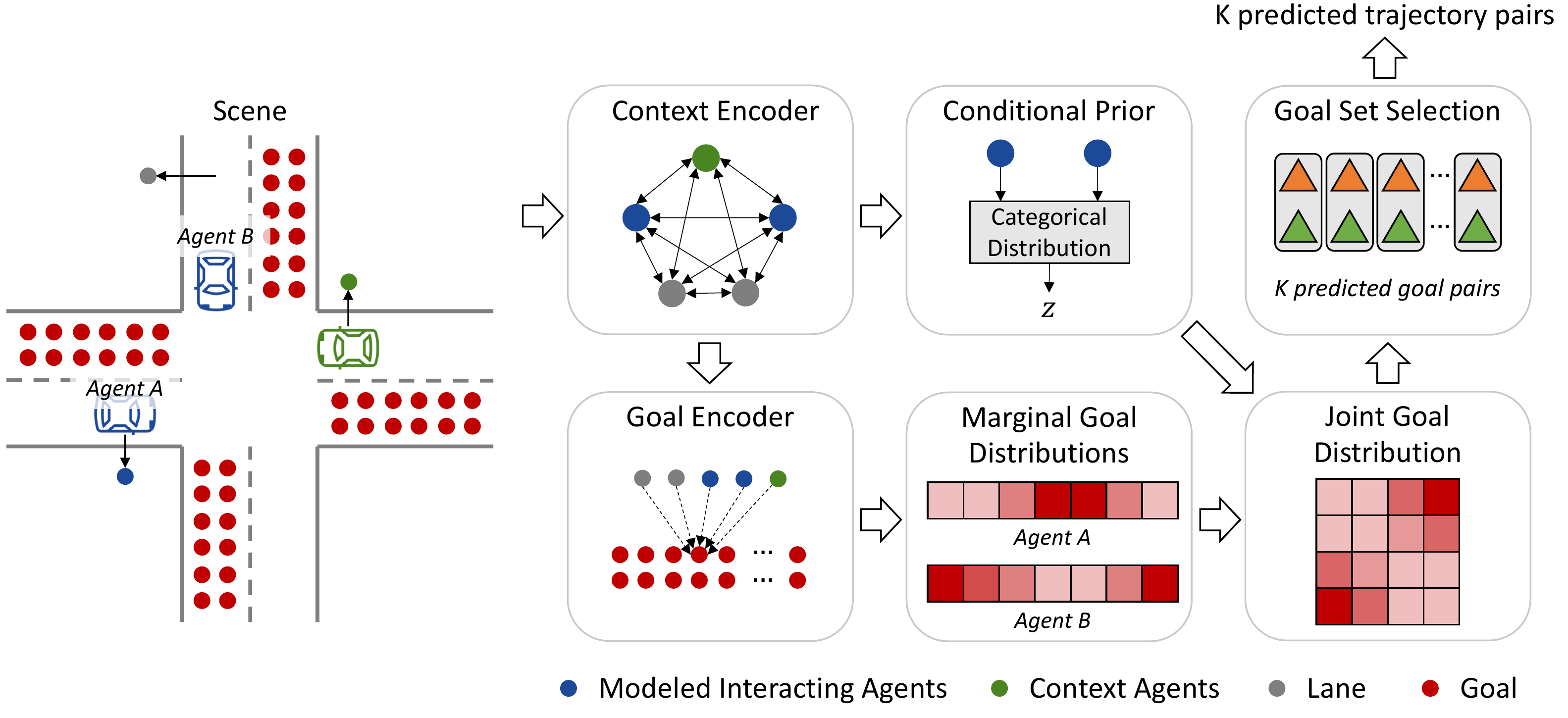}}
    \label{fig:architecture}}
    \qquad
    \subfloat[\centering Pseudo Interaction Labels]{{\includegraphics[width=2.8in]{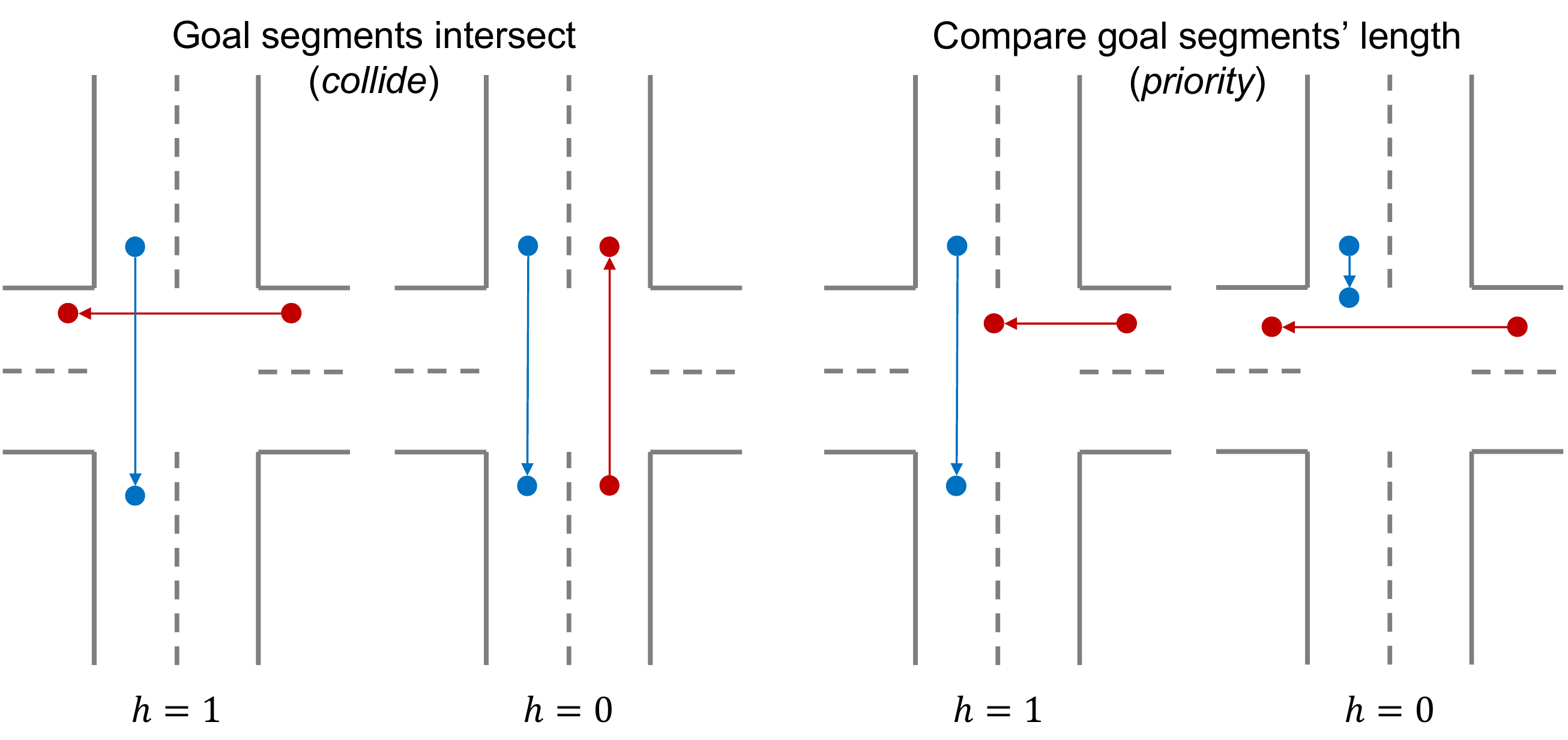}} \label{fig:pseudo_label}
    }
    \caption{Overall Model Architecture and Pseudo Interaction Labels.}
    \vspace{-0.1in}
\end{figure*}

In this section, we introduce the architecture of the model we propose for interactive trajectory prediction. As illustrated in Fig.~\ref{fig:architecture}, the model consists of three modules: 1) A marginal goal prediction module which predicts the goal distribution of each interacting agent separately; 2) A joint goal prediction module which explicitly models the joint distribution of goal pairs based on the predicted marginal distributions; 3) A trajectory completion module which predicts the trajectory of each agent conditioned on sampled goal points. 

\subsection{Modules}
\subsubsection{Marginal Goal Prediction} \label{sec:marginal-goal}
We choose DenseTNT \cite{gu2021densetnt_iccv} as the backbone model when designing the marginal goal prediction module. Specifically, we extract features of maps and agents using the vectorized encoding method proposed in \cite{gao2020vectornet}. Afterwards, we use the context embeddings to generate goal embeddings for a dense goal set $\mathcal{G}$. The goal set is sampled from the HD maps to cover the drivable area of the modeled agents. We follow DenseTNT to use the attention mechanism in~\cite{vaswani2017attention} to extract local information between the goals and the scene. We denote the embeddings obtained at this stage for the dense goals and interacting agents as $\boldsymbol{F}\in \mathbb{R}^{|\mathcal{G}|\times d_g}$ and $\boldsymbol{L}\in\mathbb{R}^{2\times d_v}$ respectively, where $d_g$ and $d_v$ are the dimensions of goal and agent embeddings. 

The interaction prediction track of WOMD has a prediction horizon of 8s. It is difficult to capture the multimodality in long-term trajectory distribution with a single goal point. We follow~\cite{gu2021densetnt} to model the goal distributions in an autoregressive manner, at 3s, 5s and 8s, respectively. To encourage the usage of interaction information in goal prediction, we add a MLP to update the interacting agents' embeddings at each timestep as follows: 
\begin{equation*}
    \boldsymbol{\hat{L}}_{t,i} = \mathrm{MLP}\left(\boldsymbol{L}_i, \boldsymbol{L}_{-i}, \boldsymbol{F}_{k^i_{1:t-1}}, \boldsymbol{F}_{k^{-i}_{1:t-1}}\right), 
\end{equation*}
where $\boldsymbol{F}_{k^i_{1:t-1}}$ collects the embeddings of the $i^{th}$ agent's goals at prior timesteps. The marginal probability of the $k^{th}$ goal for the $i^{th}$ agent at each timestep is then predicted as:
\begin{equation}
    \phi_{t,k}^i=\frac{\exp\left(\mathrm{MLP}(\boldsymbol{F}_k, \boldsymbol{\hat{L}}_{t,i})\right)}{\sum_{j=1}^{|\mathcal{G}|}\exp\left(\mathrm{MLP}(\boldsymbol{F}_j, \boldsymbol{\hat{L}}_{t,i})\right)}. \label{eqn:score-fn}
\end{equation}
At the training stage, we follow the well-known practice in autoregressive model training by feeding the ground-truth goals of the previous timesteps. 

\subsubsection{Joint Goal Prediction}
With the marginal goal distributions at timestep $t$, we first select the top-$M$ goal candidates for each agent based on their marginal probabilities and then models the joint distribution over the $M^2$ goal pair candidates. As mentioned in Sec. \ref{sec:toy_example}, we model the joint distribution with a CVAE and utilize the pseudo labels to induce an interpretable interactive latent space. The conditional prior encoder models the distribution of $\boldsymbol{z}$ conditioned on $\boldsymbol{L}$. The posterior encoder further conditions $\boldsymbol{z}$ on $\boldsymbol{F}_{k^1_{1:T}}$ and $\boldsymbol{F}_{k^2_{1:T}}$, i.e., the embeddings of the two agents' ground-truth goals. Both the conditional prior and posterior encoders are modeled with simple MLPs. To decode the joint goal distribution from a sampled $\boldsymbol{z}$, we first obtain a joint agent embedding $\boldsymbol{\Tilde{L}}_t\in \mathbb{R}^{1\times {d}_h}$ as follows: 
\begin{equation*}
    \boldsymbol{\Tilde{L}}_t = \mathrm{MLP}\left(\boldsymbol{L}, \boldsymbol{F}_{k^1_{1:t-1}}, \boldsymbol{F}_{k^2_{1:t-1}}, \boldsymbol{z}\right).
\end{equation*}
We obtain the features of goal pairs by concatenating the corresponding goals' embeddings and their marginal probabilities and then encoding into embeddings of the same dimension as the joint agent embedding with simple MLPs. We denote the resulting goal pair embeddings as $\Tilde{\boldsymbol{F}}_t\in \mathbb{R}^{M^2 \times {d}_h}$. We then use attention mechanism to gather the local information of goal pairs:
\begin{align*}
    \boldsymbol{Q}_t &= \Tilde{\boldsymbol{F}}_t\boldsymbol{W}^Q,\\
    \boldsymbol{K}_t &= \left[\Tilde{\boldsymbol{F}}_t\boldsymbol{W}_m^K; \Tilde{\boldsymbol{L}}_t\boldsymbol{W}_v^K\right],\\
    \boldsymbol{V}_t &= \left[\Tilde{\boldsymbol{F}}_t\boldsymbol{W}_m^V; \Tilde{\boldsymbol{L}}_t\boldsymbol{W}_v^V\right],\\
    \boldsymbol{\bar{F}}_t &= \mathrm{softmax}\left(\frac{\boldsymbol{Q}_t \boldsymbol{K}_t^\intercal}{\sqrt{d_k}}\right)\boldsymbol{V}_t,
\end{align*}
where $\boldsymbol{W}^Q, \boldsymbol{W}_m^K, \boldsymbol{W}_v^K, \boldsymbol{W}_m^V, \boldsymbol{W}_v^V\in \mathbb{R}^{d_h \times d_k}$ are matrices for linear projection, $d_k$ is the dimension of query / key / value vectors. We predict the joint probability of the $k^{th}$ goal pair at the given timestep in the similar way as Eqn.~\ref{eqn:score-fn}.

\subsubsection{Trajectory Completion}
The trajectory completion is similar to the one in~\cite{zhao2020tnt} and~\cite{gu2021densetnt_iccv}. Given a sequence of goals, we pass their embeddings to a simple MLP to decode the whole trajectory. The trajectories for the two agents are decoded separately. At the training stage, teacher forcing technique is applied by feeding the ground-truth goal sequences when training the trajectory completion module. 

\subsection{Training Scheme}
\label{sec:training_scheme}
To train the overall model, we first train the marginal goal prediction module together with the trajectory completion module. The loss function is the same as in~\cite{gu2021densetnt_iccv}. Afterwards, we freeze the parameters of these modules and train the joint prediction module. The objective function is essentially ELBO, but with the auxiliary losses corresponding to the three types of pseudo labels introduced in Sec. \ref{sec:toy_example}. In particular, the pseudo interaction labels are defined for each pair of segments connecting goal points at neighboring timesteps (e.g., 0s-3s, 3s-5s, 5s-8s). As illustrated in Fig.~\ref{fig:pseudo_label}, for each pair of segments, the pseudo interaction labels are two indicators showing: 1) if the goal segments of the two vehicles intersect; 2) if the goal segment of the first vehicle is longer than the one of the second vehicle. The first feature gives us a hint on whether the two vehicles have a conflict zone along their driving directions. The second feature provides a necessary condition on their right-of-way. If a vehicle has the right-of-way, it should have a larger average speed than the vehicle yielding to it. 

\subsection{Goal Selection}
At test time, we need to select a final small number of goal pairs for prediction. The most widely used algorithm is NMS. However, such a heuristic approach is difficult to tune and is not guaranteed to find the optimal solution. To address this issue, an optimization-based approach is proposed in \cite{gu2021densetnt_iccv} to select a goal set from a predicted distribution. While we may adopt it to select goal pairs at a single timestep, it still remains heuristic when sampling from the latent space as well as the autoregressive model. To ensure a fair comparison among the different model variants studied in Sec. \ref{sec:exp}, inspired by \cite{deo2020trajectory}, we instead first randomly sample $N$ sequences of goal pairs, and then fit them to a Gaussian mixture model (GMM) with $K$ components. We take the mean values of the components as the final $K$ goal pair sequences, and set the likelihood of each predicted goal pair sequence as the probability of the corresponding component.

\section{Experiments} \label{sec:exp}
We evaluate the proposed prediction model on WOMD. In particular, we focus on the interaction prediction track, where the future trajectories of an interacting pair for the next 8 seconds are predicted, given the historical observation for the past 1 second. We used the subset of the dataset with labeled interaction pairs of \emph{vehicles} for training and evaluation. With the experiments, we would like to answer: 
\begin{itemize}
    \item Do the pseudo labels induce a meaningful latent space distinguishing different interactive behaviors?
    \item Does a meaningful latent space improve prediction performance and sampling efficiency?
\end{itemize}

\textbf{Model Variants.}
Our experiment mainly focuses on ablation studies, comparing our model against multiple variants of it. We compare the performance of three models: 1) The \emph{Joint-Vanilla} model, which is our joint prediction model without the pseudo labels; 2) The \emph{Joint-NonInteract} model, which uses pseudo distance and marginal labels in addition to the vanilla version; 3) The \emph{Joint-Full} model, which is the one we propose, i.e., the joint prediction model with the auxiliary losses corresponding to all the proposed pseudo labels (i.e., distance, marginal, interaction). We do not experiment with other methods from the literature since our core contribution lies in utilizing the novel pseudo labels to induce an non-trivial and interpretable latent space. Achieving state-of-the-art performance on the benchmark is not our objective. 

\textbf{Training Settings.}
To train the overall model, we first train the marginal goal prediction module together with the trajectory completion module following most of the hyper-parameters introduced in \cite{gu2021densetnt_iccv}. Then we select $M=65$ goal candidates based on the marginal probability for each agent, and train the joint goal prediction module. We add annealing on the KL divergence weight.

\textbf{Evaluation Metrics.}
We use these metrics\textemdash minADE, minFDE, and mAP\textemdash introduced in \cite{Ettinger_2021_ICCV}, to evaluate the interactive prediction performance. The metrics for joint prediction involve the predicted trajectories of two interacting vehicles at the same time. The definitions of minADE and minFDE are similar to the single-agent case. However, the displacement errors are computed between the trajectory pairs and their ground-truth labels jointly. The mAP metric is a newly proposed metric for the Waymo Open Challenge. It computes the average precision over eight different ground-truth trajectory primitives defined based on the dataset.

\subsection{Empirical Prediction Results}
In Table~\ref{tab:metric}, we compare the prediction performance of the model variants on the validation dataset. We evaluate the prediction over 20000 validation samples in 3s, 5s, and 8s time horizons with the metrics introduced before. The results for the three time horizons are averaged and reported. We show the evaluation results based on different numbers of samples before GMM fitting, with $N=8$ and $N=120$. In all the experiments, we set $K=6$ regardless of the values of $N$ to ensure a fair comparison in prediction errors. From Table~\ref{tab:metric}, we can see that the prediction performance is sensitive to the sample number $N$. With larger $N$, the sampled trajectories are more likely to cover the multimodality in joint distribution, which leads to more diverse and accurate prediction after GMM fitting. From the table, we can see that the \emph{Joint-Full} model always has better performance under the same sample number $N$. Note that in online prediction, the maximum allowable $N$ is directly determined by the required computational time. Our purpose is to get accurate and diverse predictions with a small sample number $N$ to enable efficient online inference. We indeed observe a larger improvement with the use of pseudo labels when $N=8$ compared to $N=120$. 

\begin{table}[t]
	\caption{Validation Results on All Samples}
	\centering
	\begin{tabular}{l| c |c |c}
	\hline
	Method & minADE & minFDE & mAP \\
	\hline  \hline

    Joint-Vanilla,        $N$=120  & 1.58 & 3.44 & 0.078 \\
    Joint-Full,  $N$=120 (Ours) & 1.55 & 3.33 & 0.084 \\
    \hline
    Joint-Vanilla         $N$=8   & 1.98 & 4.28 & 0.020 \\
	Joint-Full           $N$=8 (Ours) & \textbf{1.89} & \textbf{4.09} & \textbf{0.027}\\
	
	\hline
	\hline
	\end{tabular}
	\label{tab:metric}
\end{table}

\begin{table}[t]
	\caption{Ablation Study on Strong-Interactive Samples}
	\centering
	\begin{tabular}{l| c |c }
	\hline
	Method                         & minADE & minFDE \\
	\hline  \hline
    Joint-Vanilla,        $N$=8    & 1.89 (0.06) & 4.11 (0.17)   \\
    Joint-NonInteract,       $N$=8    & 1.88 (0.04) & 4.02 (0.07)  \\
    Joint-Full,           $N$=8 (Ours)  & \textbf{1.76} (0.02) & \textbf{3.78} (0.04)  \\
	\hline
	\hline
	\end{tabular}
	\vspace{-0.1in}
	\label{tab:sample_compare}
\end{table}

To evaluate our proposed joint prediction model in highly interactive scenarios, we select a set of strong-interactive cases from the validation dataset. Joint modeling the behavior of the interacting agents is critical for these highly interactive scenarios, which is the main motivation behind our proposed method. We select the data samples where goal segments of two vehicles intersect, by using the pseudo interaction labels introduced in Sec. \ref{sec:training_scheme}. The prediction results of models using different pseudo labels are shown in Table \ref{tab:sample_compare}. Since mAP is extremely sensitive to the hyper-parameters when $N$ is small, we do not consider the mAP comparison for quantitative analysis. As the number of selected samples is small compared to the complete validation set (351 of 20000), we evaluate each model three times and report the mean and the standard deviation. We observe a significant improvement in prediction performance and stability by adding interaction pseudo labels (\emph{Joint-Full} model). With a well-trained latent space, we are more likely to cover more interaction patterns even if the number of samples is limited, leading to smaller prediction errors in these strong-interactive cases, especially when $N$ is small. 

\begin{figure*}[t]
    \centering
    \subfloat[Different speed]{{\includegraphics[height=2.7cm]{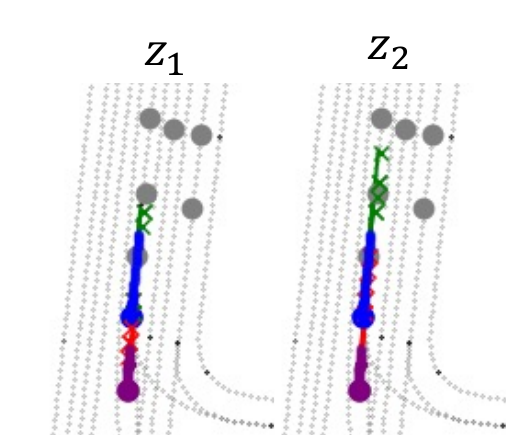} }\label{fig:latent_sample_example_1}}
    \qquad
    \subfloat[Different right-of-way]{{\includegraphics[height=2.7cm]{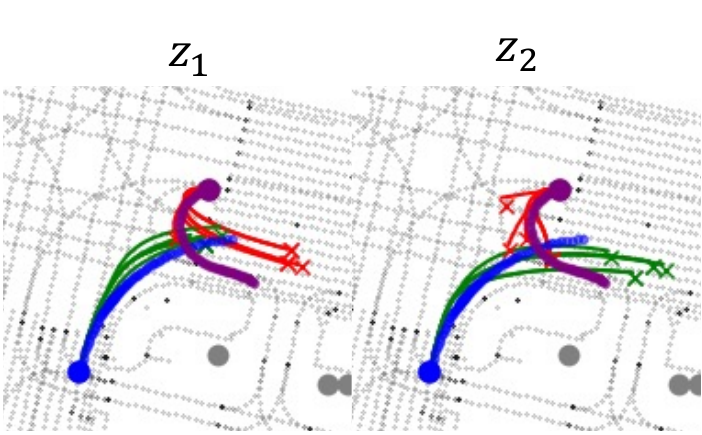} \label{fig:latent_sample_example_2}}}
    \qquad
    \subfloat[Different route selection]{{\includegraphics[height=2.6cm]{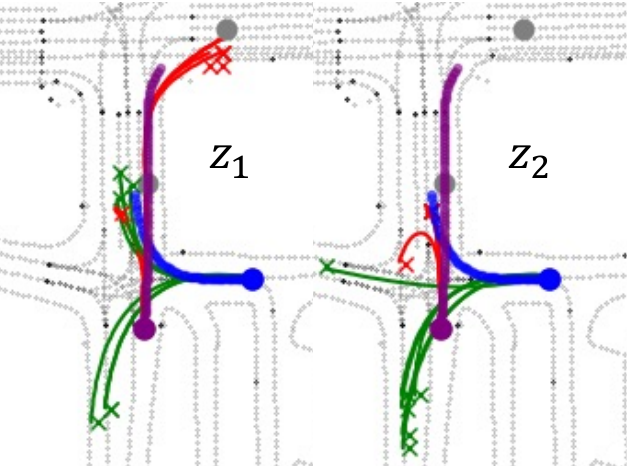} \label{fig:latent_sample_example_3}}}
    \subfloat{{\includegraphics[height=2.6cm]{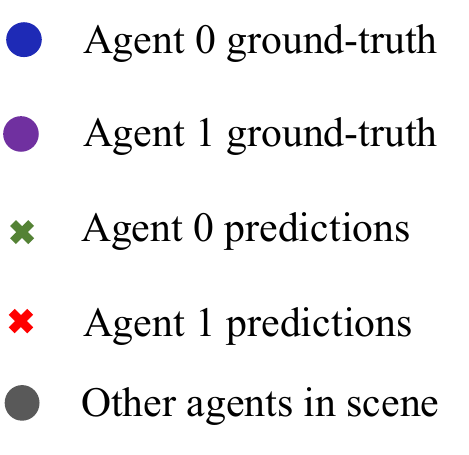}}}
    \caption{Comparison of 6 sampled first-step goal predictions conditioned on 2 different selected latent $z$ value \emph{Joint-Full}. Different interaction modes can be found in different latent values, meaning we have learned a meaningful latent space.}
    \vspace{-0.1in}
    \label{fig:latent_sample}
\end{figure*}

\subsection{Latent Space learned by CVAE with Pseudo Labels}
During training, we indeed observed that pseudo labels, especially marginal pseudo labels, help avoid KL vanishing in most cases. To demonstrate the interactive pattern encoded by latent space, we visualize predicted trajectories for selected interactive scenarios from the dataset, as shown in Fig.~\ref{fig:latent_sample}. We use the \emph{Joint-Full} model under different latent variables in the same scenario to make these predictions. Given the historical information, we sample six different goal pairs from the joint goal distribution prediction model conditioning on two different discrete latent variable $z$ with the largest probabilities. In Fig.~\ref{fig:latent_sample}, we can clearly see two different interaction modes with different $z$. The agents either change their speed, route, right-of-way or combinations of these features when switching the latent variables. Meanwhile, the \emph{Joint-Vanilla} model fails to give a separated latent space (e.g., predictions sampled from different latent variables are similar) in the same scenarios, because of KL vanishing. This shows that our proposed model indeed learns an interpretable latent space capturing the interaction modes inherited from the pseudo labels. 

\section{Conclusion}
In this work, we study the interaction prediction problem under the goal-conditioned framework. To develop an interpretable and sampling-efficient prediction model, we leverage the CVAE framework to explicitly capture diverse interaction modes in joint goal distribution. We find the vanilla model is prone to suffering from posterior collapse, resulting in a totally uninformative latent space. We explore the underlying reasons in a toy example, and propose a general and flexible approach to mitigate this issue with pseudo labels incorporating domain knowledge on interaction. We show that the pseudo labels guide the model to learn an interpretable latent space in our experiments. 

\bibliographystyle{ieeetr} 
\bibliography{IEEEabrv,reference}

\clearpage

\end{document}